\documentclass{article} %
\pdfoutput=1

\usepackage{iclr2016_conference,times}

\usepackage{epsfig}
\usepackage{graphicx}
\usepackage{amsmath}
\usepackage{amssymb}
\usepackage{booktabs}
\usepackage{multirow,array}
\usepackage{color}
\usepackage{bm}
\usepackage{subcaption}
\usepackage{ifthen}
\usepackage[font=footnotesize,subrefformat=parens]{subcaption}
\usepackage{tabu}           %
\usepackage{xcolor}            %
\usepackage{soul}
\usepackage{tikz}
\usetikzlibrary{arrows,shapes,positioning}
\usetikzlibrary{calc,decorations.markings}

\usepackage[pagebackref=true,breaklinks=true,letterpaper=true,colorlinks,bookmarks=false]{hyperref}
\usepackage{url}

\definecolor{Plum}{RGB}{142, 69, 133}
\definecolor{Cyan}{RGB}{0, 255, 255} %
\definecolor{Red3}{HTML}{a40000}
\definecolor{Green3}{HTML}{4e9a06} %

\definecolor{tabutter}{rgb}{0.98824, 0.91373, 0.30980}      %
\definecolor{ta2butter}{rgb}{0.92941, 0.83137, 0}     %
\definecolor{ta3butter}{rgb}{0.76863, 0.62745, 0}     %

\definecolor{taorange}{rgb}{0.98824, 0.68627, 0.24314}      %
\definecolor{ta2orange}{rgb}{0.96078, 0.47451, 0}     %
\definecolor{ta3orange}{rgb}{0.80784, 0.36078, 0}     %

\definecolor{tachocolate}{rgb}{0.91373, 0.72549, 0.43137}   %
\definecolor{ta2chocolate}{rgb}{0.75686, 0.49020, 0.066667} %
\definecolor{ta3chocolate}{rgb}{0.56078, 0.34902, 0.0078431}   %

\definecolor{tachameleon}{rgb}{0.54118, 0.88627, 0.20392}   %
\definecolor{ta2chameleon}{rgb}{0.45098, 0.82353, 0.086275} %
\definecolor{ta3chameleon}{rgb}{0.30588, 0.60392, 0.023529} %

\definecolor{taskyblue}{rgb}{0.44706, 0.56078, 0.81176}     %
\definecolor{ta2skyblue}{rgb}{0.20392, 0.39608, 0.64314} %
\definecolor{ta3skyblue}{rgb}{0.12549, 0.29020, 0.52941} %

\definecolor{taplum}{rgb}{0.67843, 0.49804, 0.65882}     %
\definecolor{ta2plum}{rgb}{0.45882, 0.31373, 0.48235}    %
\definecolor{ta3plum}{rgb}{0.36078, 0.20784, 0.4}     %

\definecolor{tascarletred}{rgb}{0.93725, 0.16078, 0.16078}  %
\definecolor{ta2scarletred}{rgb}{0.8, 0, 0}        %
\definecolor{ta3scarletred}{rgb}{0.64314, 0, 0}       %

\definecolor{taaluminium}{rgb}{0.93333, 0.93333, 0.92549}   %
\definecolor{ta2aluminium}{rgb}{0.82745, 0.84314, 0.81176}  %
\definecolor{ta3aluminium}{rgb}{0.72941, 0.74118, 0.71373}  %

\definecolor{tagray}{rgb}{0.53333, 0.54118, 0.52157}     %
\definecolor{ta2gray}{rgb}{0.33333, 0.34118, 0.32549}    %
\definecolor{ta3gray}{rgb}{0.18039, 0.20392, 0.21176}    %

\newcommand{\reffig}[1]{Figure~\ref{fig:#1}}
\newcommand{\refsec}[1]{Section~\ref{sec:#1}}

\newcommand{\reftbl}[1]{Table~\ref{tbl:#1}}

\newcommand{\refeq}[1]{Equation~\eqref{eq:#1}}

\newcommand{\lblfig}[1]{\label{fig:#1}}
\newcommand{\lblsec}[1]{\label{sec:#1}}
\newcommand{\lbleq}[1]{\label{eq:#1}}
\newcommand{\lbltbl}[1]{\label{tbl:#1}}

\definecolor{lightpink}{rgb}{1.0, 0.71, 0.76}
\definecolor{wildwatermelon}{rgb}{0.99, 0.42, 0.52}

\title{Constrained Structured Regression with\\Convolutional Neural Networks}

\author{Deepak Pathak, Philipp Kr\"ahenb\"uhl, Stella X. Yu \& Trevor Darrell\\
Department of Computer Science\\
UC Berkeley\\
\texttt{\{pathak,philkr,stellayu,trevor\}@berkeley.edu}
}

\begin{document}

\maketitle

\begin{abstract}
Convolutional Neural Networks (CNNs) have recently emerged as the dominant model in computer vision.
If provided with enough training data, they predict almost any visual quantity.
In a discrete setting, such as classification, CNNs are not only able to predict a label but often predict a confidence in the form of a probability distribution over the output space.
In continuous regression tasks, such a probability estimate is often lacking.
We present a regression framework which models the output distribution of neural networks.
This output distribution allows us to infer the most likely labeling following a set of physical or modeling constraints.
These constraints capture the intricate interplay between different input and output variables, and complement the output of a CNN.
However, they may not hold everywhere.
Our setup further allows to learn a confidence with which a constraint holds, in the form of a distribution of the constrain satisfaction.
We evaluate our approach on the problem of intrinsic image decomposition, and show that constrained structured regression significantly increases the state-of-the-art.
\end{abstract}

\section{Introduction}
Structured regression lies at the heart of some of the most active computer vision problems of our time.
Examples include optical flow~\citep{horn1981determining}, monocular depth estimation~\citep{eigen2014depth}, intrinsic image decomposition~\citep{barronTPAMI2015} etc.
Convolutional neural networks (CNNs)~\citep{fukushima1980neocognitron,krizhevsky2012imagenet,lecun1989backpropagation} have greatly advanced the state of the art in all those structured output tasks \citep{eigen2014depth,yu:albedo15}.
However CNNs predict each output independently, and thus ignore the intricate interplay of the output variables imposed by physical or modeling constraints.
Instead they are forced to learn all physical properties of a scene directly from the training data, and often fail due to the limited capacity of the model.

In this work, we propose to bring those dependencies back to deep structured regression, in the form of constraints on the output space.
A constraint ties the output of several regression targets together.
In a naive first approach, we learn a standard deep structured regression model and find the closest solution to the predicted structured output that follows the constraints strictly, using a simple Euclidean distance measure.
This results in an averaging of the output.
It makes only limited use of the training data, and further assumes that the constraints are always satisfied, which is not true in general.
For instance, in the task of intrinsic images, Lambert's law~\citep{barrow1978computer} assumes that surfaces have diffused reflectance, which means product of shading and albedo images is equal to the original image.
However, this is not true for specular surfaces like mirrors, metals etc~\citep{zhang1999shape}, such that a naive approach does not perform well in those areas.

To make full use of the training data, we regress not just to a single output variable, but rather a fully factorized distribution over possible outputs.
We further predict a distribution over each of our constrains, allowing it not be violated under certain conditions.
These distributions capture a the confidence with which the model makes its predictions, or the confidence that a certain constraint holds.
A highly confident prediction is peaked around the correct answer, while an uncertain prediction will be more uniform.
We use these confidences, and pick the most likely output labeling following our constraints.
This allows the model to trust outputs differently during inference, see \reffig{methodFig} for an overview of our framework.

We apply our structured regression framework to the problem of intrinsic image decomposition~\citep{barrow1978computer}.
The goal of intrinsic image decomposition is to decompose the input image into albedo (also called reflectance image) and shading images.
The output space, in such tasks, has dependencies on the input which can be modeled as physics based constraints such as Lambertian lighting assumption in intrinsic image decomposition~\citep{barrow1978computer}.
At inference time, we find a structured output following those constraints.
This alleviates the pressure on the CNN to explicitly learn all physical properties of a scene, and allows it to focus more on the statistical correlations between the input image and the output.

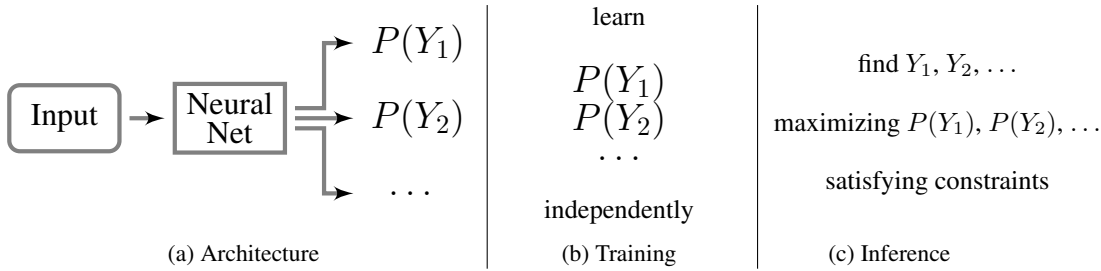
\begin{figure}[t]
\centering

  \tikzset{
    block/.style={rectangle, fill=white, draw=gray, ultra thick, text centered},
    Ys/.style={minimum height=2.5em, text width = 3em, text centered},
    line/.style={draw=gray, ultra thick, -latex', shorten >= 3pt, shorten <= 3pt},
    -|-/.style={
      to path={
        (\tikztostart) -| ($(\tikztostart)!#1!(\tikztotarget)$) |- (\tikztotarget)
        \tikztonodes
      }
    },
    -|-/.default=0.3,
    }
 \begin{subfigure}[b]{.45\textwidth}
   \begin{tikzpicture}
    \node[block,rounded corners, text width=3.5em, minimum height=2.5em] (input) {\large Input};
    \node[block, minimum height = 2.5em, text width=3.5em, right of=input, node distance=2.2cm] (nn) {\large Neural Net};
    \draw [line] (input) -- (nn);
    \node[Ys, right of=nn, node distance=2.4cm] (y2) {\Large $P(Y_2)$};
    \node[Ys, above of=y2, node distance=1cm] (y1) {\Large $P(Y_1)$};
    \node[Ys, below of=y2, node distance=1cm] (yn) {\Large $\ldots$};
    \draw [line] (nn.10) to[-|-] (y1);
    \draw [line] (nn.0) to[-|-] (y2);
    \draw [line] (nn.350) to[-|-] (yn);
   \end{tikzpicture}
  \caption{Architecture}
 \end{subfigure}
 \vline
 \begin{subfigure}[b]{.25\textwidth}
  \centering
  \begin{tabular}{c}
   learn\\
   \\
   \Large $P(Y_1)$\\
   \Large $P(Y_2)$\\
   \Large $\ldots$\\
   \\
   independently
  \end{tabular}

  \caption{Training}
 \end{subfigure}
 \vline
 \begin{subfigure}[b]{.25\textwidth}
  \begin{tabular}{c}
   find $Y_1$, $Y_2$, $\ldots$\\
   \\
   maximizing $P(Y_1)$, $P(Y_2)$, $\ldots$\\
   \\
   satisfying constraints
   \\
   \\
  \end{tabular}
  \caption{Inference}
 \end{subfigure}
\caption{Illustration of our constraint regression framework. We learn the distribution of outputs as well as constraints at training time. At inference, we sample from the learned distribution of output such that the constraints are satisfied.}
\lblfig{methodFig}
\end{figure}

In summary, our constrained regression framework learns not only to capture the ground-truth values, but also to capture the variation or confidence in its own predictions.
Moreover, our constraints re-introduce the coupling between albedo and shading that has been ignored in the prior supervised learning.
We achieve significant improvement over the state of the art performance
and show large visual improvement on MPI Sintel dataset.

\section{Related Work}
\paragraph{Modeling Weight Uncertainty in Neural Networks}
In neural networks, uncertainty is usually modeled as distributions over the learned network weights~\citep{denker1991transforming,mackay1992practical}.
If the weights are drawn from a Gaussian distribution then the neural network approximates a Gaussian process in the limit.
In such a network, the output distribution is inferred through Bayesian inference.
Recently, \citet{blundell2015weight} presented a back-propagation based variational technique to learn weight uncertainty.
The Bayesian inference based regularization in deep networks is closely related to dropout~\citep{Gal2015DropoutB}.
Our approach in this paper is not to learn weight uncertainty, but to directly learn the parameters by assuming some output distribution.
This allows for much faster inference and training by leveraging the standard feed forward inference and discriminative training for network parameters.

\paragraph{Deep Structured Prediction}
Most of the problems involve predicting multiple outputs which may be dependent. 
Such structured prediction tasks are mostly modeled using MRFs.
To extend the success of CNN on a small output space to a large one, if the global scoring function on the output can be decomposed into the sum of local scoring functions depending on a small subsets of the output, \citet{chen2014learning} shows that the loss function is still representable and there could be an efficient iterative solution.
\citet{jaderberg2014deep} train CNN to model unaries for the structured prediction task in a max-margin setting for text recognition.
Jointly learning CRF and CNN using mean field inference has been shown to achieve great results in segmentation\citep{zheng2015conditional}.
However, most of these works are applied for discrete prediction tasks.
For continuous settings, great results have been achieved in depth estimation using CRF
formulations, where unary potentials come from a CNN that
maps the image to the depth value at single
pixels or super-pixels, and where pairwise potentials reflect the
desired smoothness of depth predictions between neighboring pixels or
segments \citep{Liu_2015_CVPR, Wang_2015_CVPR,Li_2015_CVPR}.
Compared to these models, not only our CNN formulation captures the variation in its own predictions, but our constraints also provide an easier and more efficient alternative to higher-order clique potentials.
We view structured regression task as constraint satisfaction problem and show that learning variance and distribution of output makes it possible to learn and enforce constraints at inference.

\paragraph{Intrinsic Image Decomposition}
Intrinsic image decomposition by definition is ambiguous.
Traditional approaches seek various physics or statistics based
priors on albedo and shading such as sparse and piecewise constant
albedo and smooth shading ~\citep{horn:retinex74,grosse2009ground},
or more recently from additional depth images
\citep{lee2012estimation,barronTPAMI2015,chen2013simple}.
MIT intrinsics, Sintel, and Intrinsics In the Wild datasets~\citep{grosse2009ground,sintel,bell2014intrinsic} provide ground-truth for
intrinsics and open up new directions with data-driven approaches
that have largely got rid of any hand-designed features and complex
priors \citep{yu:light15,ZKE:ICCV:2015,zoran_2015_ICCV,yu:albedo15}.
\citet{yu:light15} regress to a globally consistent ranking of albedo values, however they do not obtain a full intrinsic image decomposition.
\citet{ZKE:ICCV:2015} and \citet{zoran_2015_ICCV} both predict the pairwise ordering of albedo values and then a CRF or a constrained quadratic program to turn this pairwise orderings into a globally consistent and spatially smooth intrinsic image decomposition.
Neither method attempted to predict the intrinsic image decomposition directly, due to the lack of training data for real world images.
Direct intrinsics \citep{yu:albedo15} is the first brute-force
approach aiming to associate albedo and shading directly with the
input image by training convolutional regressors from synthetic
ground-truth data.  However, lost in the simplicity of regression
are the strong coupling constraints in the form of the intrinsics decomposition
equation.  Our formulation addresses this weakness.

\section{Preliminaries}
\lblsec{prelim}
We first introduce the necessary notation and define the common learning objective for structured regression task.
We then give a probabilistic interpretation of this regression.
\refsec{technical} combines this structured regression with a set of physical inference-time constraints, in our constrained structured regression framework.
We learn a complete probabilistic model for structured regression, modeling uncertainties in both the output variables and the constraints.
This derivation is general and not application specific.
\refsec{intrinsics} then applies our structured regression to intrinsic image decomposition.
\refsec{details} provides more details on our network architecture and training procedure.

Consider the problem of regression to an output $Y \in \mathrm{R}^N$ given an input image $I$.
A deep convolutional regression approximates $Y = f(I;\theta)$ with a multi-layer convolutional neural network with parameters $\theta$.
Those parameters $\theta$ are learned from a set of training images $\mathcal{I} = \{I_1,I_2,\ldots\}$ and corresponding ground truth targets $\mathcal{\tilde Y} = \{\tilde Y_1, \tilde Y_2, \ldots\}$.
One of the most common learning objectives is a Euclidean loss :
\begin{equation}
 \ell(I, \tilde Y) = (f(I;\theta) - \tilde Y)^2,\lbleq{l2}
\end{equation}
for each image $I \in \mathcal{I}$ and target $\tilde Y \in \mathcal{\tilde Y}$ in our training set.
This Euclidean loss has a nice probabilistic interpretation as the negative log-likelihood of a fully factorized normal distribution $Y_k \sim \mathcal{N}(\mu_k,\sigma_k^2)$ with mean $\mu_k=f_k(I;\theta)$ and unit standard deviation $\sigma_k=1$ for each output $Y_k$~\citep{koller2009probabilistic}.
In the following section, we use this probabilistic interpretation to enforce constrains on the output space, and find the most likely output $Y$ following these constraints.

\section{Constrained Structured Regression}
\lblsec{technical}
Many structured tasks have an inter-dependency of input and output spaces, which may be known either due to physical properties of the problem or due to a modeling choice.
Let's suppose that this knowledge is captured by the following family of constraints:
\begin{align}
\mathcal{G}(Y,I)=0 \lbleq{constraint},
\end{align}
where $\mathcal{G}$ is a constraint function acting on the output $Y$ and the input image $I$.

In practice, we limit ourselves to affine constraints $\mathcal{G}(Y,I) = A Y + B I + c$, which results in an efficient inference due to the convexity of constrained region.
Note that such a property, although not necessary, is desirable and thus informs our modeling choice.
For instance, we perform regression in log domain in intrinsics so that constraints are affine.
See Sections \ref{sec:intrinsics} for more details.

\subsection{Constrained inference}
During inference, we want to find the most likely output estimate $Y$ under the probability distribution predicted by the CNN such that it follows our known family of constraints.
Recall from \refsec{prelim} that all outputs are modeled by a normal distribution $Y \sim \mathcal{N}(f(I;\theta), 1)$, where the mean is learned.
Lets denote the probability of an output $Y$ as $P_\mathcal{N}(Y|\theta) = \mathcal{N}(Y; f(I;\theta), 1)$.
Thus, finding the most likely output $Y$ can be written as a constrained optimization problem, at inference time, as follows:
\begin{align}
\underset{Y}{\text{minimize}}\quad &- \log P_\mathcal{N}(Y|\theta) = (Y - f(I; \theta))^2 \notag \\
\text{subject to}\quad &\mathcal{G}(Y,I)= 0 .\lbleq{inference}
\end{align}
If constraints are affine in the output $Y$, this inference is convex for a fixed set of training parameters.
Specifically, this is a quadratic program because the negative log likelihood $-\log P_\mathcal{N}$ is a quadratic function, and thus can be solved either in closed form or using QP solvers.

Note that in \refeq{inference}, all outputs $Y_k$ are assumed to have the same distribution with different mean values.
This is obviously not true, as some quantities are easier to estimate than others, hence the CNN should be more confident in some areas than others.
We will now show how to learn a more general capturing a full distribution for each output $Y_k$.

\subsection{Probabilistic regression}
\lblsec{prob_reg}
We use the probabilistic interpretation of the Euclidean loss (\ref{eq:l2}) and also learn the standard deviation of our output as  $\sigma_k^2=g_k(I;\theta)$.
This standard deviation can be interpreted as a simple confidence estimate of our prediction.
A small standard deviation signifies a confident prediction, while a large standard deviation corresponds to an uncertain prediction.
The resulting learning objective is expressed as follows:
\begin{align}
\ell(I, \tilde Y) &= -\sum_k \log P_\mathcal{N}(\tilde Y_k | \mu_k,\sigma_k^2) \notag\\
&= \frac{1}{2} \sum_k\left(\frac{(f_k(I;\theta)-\tilde Y)^2}{g_k(I;\theta)} + \log g_k(I;\theta) + \log 2 \pi \right) \lbleq{lossEq},
\end{align}
where $P_\mathcal{N}$ is a normal distribution, and both $f$ and $g$ are convolutional neural networks parametrized by $\theta$, learned through the same objective.
This is standard negative log-likelihood minimization~\citep{koller2009probabilistic}.
For a fixed $g$ this reduces to Euclidean loss.
On the other hand, if $g$ is completely free i.e. no regularization, then the objective reduces to an L1 norm with $g_k = |f_k-\tilde Y|$ predicting the magnitude of that norm, hence the predicted error.
In practice, we parametrize $g$ as the exponential of the network output to ensure positivity.
We also add a small l2 regularization to the objective~\eqref{eq:lossEq}.

This probabilistic regression allows us to reason about which outputs to follow more strictly when enforcing our constraints.
However all constraints are still strictly enforced.
This is fine for setups where it is known that the constraint have to be satisfied, but it is not true for a general scenario as all constraints might not hold for all outputs.
Motivated by this, we now consider to learn the uncertainty in constraints as well and then incorporate it in the inference time optimization, similar to~\refeq{inference}.

\subsection{Probabilistic constraints}
\lblsec{prob_constraint}
In structured prediction tasks, we have the knowledge of constraints that our output should satisfy.
If the constraints are not strictly satisfied everywhere, the easiest mechanism would be to learn the distribution over constraint satisfaction in a similar way as we learned the distribution of output in our structured regression setup.
More specifically, for $i^{th}$ constraint we write:
\begin{align}
\mathcal{G}_i(Y,I) \sim \mathcal{H}(0,\sigma_i^2) \lbleq{prob_constraint},
\end{align}
where $\mathcal{H}$ is a continuous probability distribution.
In our experiments, we restrict $\mathcal{H}$ to be a zero-mean Gaussian or Laplace distribution with learned standard deviation, $\sigma_i^2 = h_i(I;\theta)$, where $h$ is a  convolutional neural network with same parameters $\theta$ as for outputs.
For a given output, if the standard deviation $\sigma_i$ is low, then our confidence about the $i^{th}$ constraint being satisfied will be high and vice-versa.

To learn this distribution $\mathcal{H}$, we again follow the standard negative log-likelihood minimization, i.e. $\underset{\mathcal{G}}{\text{minimize}} \sum_i \left( -\log P_{\mathcal{H}}(\mathcal{G}_i | 0,\sigma_i^2) \right)$, as mentioned in detail in~\refsec{prob_reg}. 
We now see how can we incorporate this constraint modeling in the inference framework.

\subsection{Probabilistic constrained inference}
\lblsec{prob_inference}
We conclude by combining the modeled distributions of outputs and constraints in a joint optimization problem to obtain a better estimate of output.
Incorporating distributions of output in the optimization is similar to the one described in~\refeq{inference}, however, handling the distribution of constraints is not apparently obvious.
To address this, interestingly, we can express \refeq{prob_constraint} in terms of a slack variable.
This interpretation reduces $\mathcal{H}$ to a distribution over slack on the family of constraints $\mathcal{G}$.
Thus, we find most likely output $Y$ subject to certain constraints such that the slack on the constraints follows distribution $\mathcal{H}$.
This is written as follows:
\begin{align}
\underset{Y,\epsilon}{\text{minimize}}\quad & - \log P_\mathcal{N}(Y|\theta) -\log P_H(\xi|\theta) \notag \\
\text{subject to}\quad & \mathcal{G}_i(Y,I)= \xi ,\lbleq{prob_inference}
\end{align}
where $\xi$ is the slack random variable.
The resulting formulation, if constraints are affine, is again convex for a fixed training parameters.
For constraints modeled as a Gaussian or Laplacian distribution, the optimization~\eqref{eq:prob_inference} is again a Quadratic Program.
Hence, it can be optimized easily for an accurate solution.

We can better understand the~\refeq{prob_inference} in terms of confidences i.e. standard deviation.
If constraint has higher confidence, we would like to change the outputs with less confidence more and stick to the predicted value in case of outputs with more confidence to obtain the final estimates, and vice-versa.
This process is jointly achieved in this optimization since on taking log-likelihood the standard deviation becomes the weight of the squared error as discussed in~\refsec{prelim}.

\subsection{Relation to Conditional Random Fields}
So far, we have looked at the inference process from a constrained optimization perspective. 
We want to enforce some known constraints on a structured output space.
We first introduce the hard constraint scenario in~\refeq{inference}, then learn the distribution over outputs (Eq.~\ref{eq:lossEq}) and constraints (Eq.~\ref{eq:prob_constraint}) to jointly phrase the final optimization in~\refeq{prob_inference} as a constrained optimization problem with a distribution over slack. 
In this section, we discuss an alternative perspective to directly justify the~\refeq{prob_inference} by showing a connection with conditional random fields (CRF).

Lets ignore the l2 regularization for this analogy. 
We can relate the distribution defined for our output in~\refsec{prelim} to the unary potential of a CRF i.e. $\Psi_k(Y_k;\theta)=P_\mathcal{N}(Y_k|\theta)$.
Further, the constraint distribution to define the joint distribution over outputs can be re-parameterized by the constraint as  $\Psi_i(Y;\theta)=P_\mathcal{H}(\mathcal{G}_i(Y)|\theta)$.
The joint inference of such a CRF would be similar to the Langrangian dual of the problem~\eqref{eq:prob_inference}.
The log-likelihood learning formulation described in our setup can then be theoretically justified using the piecewise-learning procedure proposed by~\citet{sutton2009piecewise}.
During piecewise learning of factors, the contribution of individual outputs often gets over-counted which is determined using cross validation~\citep{shotton2009textonboost,sutton2009piecewise}.
However, we empirically found the cross-validation not to be necessary and weight the unary and pairwise terms equally, which gets handled by Langrangian.

\begin{figure*}[t]
\centering
\resizebox{\textwidth}{!}{%
\begin{tabular}{c@{\hskip 1mm}c@{\hskip 1.2mm}c@{\hskip 1.2mm}c@{\hskip 1.2mm}c@{\hskip 1.2mm}c}

\rotatebox[origin=lc]{90}{\;\;\;\;\;\;\;\;Albedo} &
\includegraphics[width=0.2\linewidth]{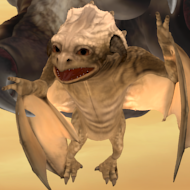} &
\includegraphics[width=0.2\linewidth]{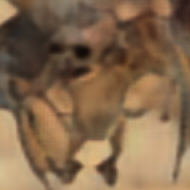} &
\includegraphics[width=0.2\linewidth]{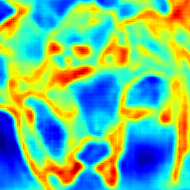} &
\includegraphics[width=0.2\linewidth]{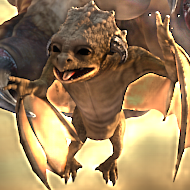} &
\includegraphics[width=0.2\linewidth]{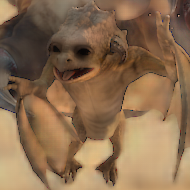}\\

\rotatebox[origin=lc]{90}{\;\;\;\;\;\;\;\;Shading} &
\includegraphics[width=0.2\linewidth]{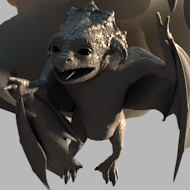} &
\includegraphics[width=0.2\linewidth]{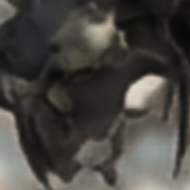} &
\includegraphics[width=0.2\linewidth]{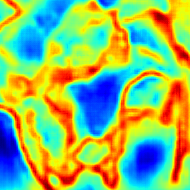} &
\includegraphics[width=0.2\linewidth]{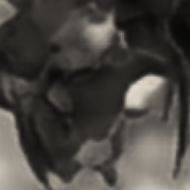} &
\includegraphics[width=0.2\linewidth]{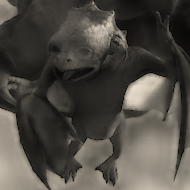}\\

&{\footnotesize Ground Truth} & {\footnotesize Direct Regression} & {\footnotesize Confidences} & {\footnotesize Hard Constraints} & {\footnotesize Probabilistic Constr.}\\
\end{tabular}}
\caption{Qualitative ablation for evaluating the contribution of our constrained structured regression pipeline. Direct regression doesn't capture fine details. If we naively enforce hard constraints, we see color artifacts due to non-lambertian surfaces. Our final probabilistic constrained output results into a detailed output with appropriate albedo.}
\lblfig{ablationFig}
\end{figure*}

\section{Intrinsic Image Decomposition}
\lblsec{intrinsics}
As a proof of concept, we apply our approach to the intrinsic image decomposition.
Given an input image $I$, the task is to decompose it into albedo $A$ and shading $S$ (also called reflectance) image.
The actual color of the contents in image is captured in albedo, while shading captures the illumination and shape information.
The physical constraint which is usually considered in this setting is based on Lambert's law, i.e. $I=A*S$. 
However, it holds only for the surfaces where incident light is diffused equally in all directions, also called Lambertian surfaces.
It is not true for Specular surfaces, like mirrors, which maintain the direction of reflection.
This constraint provides an important cue for inferring $A$ and $S$ given $I$.
Mostly, this constraint is assumed to be true everywhere and only one of $A$ or $S$ is optimized~\citep{chen2013simple,barronTPAMI2015}.
\citet{yu:albedo15} use CNN to regress to the targets $A,S$, hoping that it would implicitly learn these dependencies which is not true since the outputs are blurry and doesn't contain any high frequency information.
By reasoning and combining evidence from both albedo and shading CNN, we achieve significant improvement.

To keep the optimization convex at inference, we aim to keep the constraints affine in terms of output.
In this setting, working in log domain ensures that the intrinsic constraint is affine.
Thus our constraint is $\mathcal{G} = A + S - I = 0$, where $I$, $A$ and $S$ are the image, albedo and shading in log domain respectively.
The albedo $A$ is a three channeled RGB image.
Shading $S = B \cdot C$ is modeled by single channel gray-scale image $B$, capturing the light intensity, and global color value $C$, capturing the color of the light.
We found this formulation to be numerically more stable than modeling $S$ using three scalars per pixel.

We now discussion the instantiation of our method for this task.

\subsection*{Learning}
We learn both the distribution of outputs as well as the constraints as discussed in \refsec{technical}.
Our outputs are $Y_1=A$ and $Y_2=S$.
Their distribution is learned using the log-likelihood loss defined in~\refeq{lossEq}.
We learn the distribution of Lambertian constraint $G$ using loss described in~\refsec{prob_constraint}.
Ideally one should model $\mathcal{H}$ as Laplacian distribution, as the true violation follows it closely.
However in practice we found that Gaussian works equally well and is easier to optimize.

We tie the standard deviation of albedo distribution $\sigma_A$ across channels at each pixel.
Thus, our estimated standard deviations are single channeled heatmaps for each of the output as well as constraint.
To summarize, our CNN at training learns to predict three channeled albedo $A$, and single channeled shading $S$, $\sigma_A$, $\sigma_S$, $\sigma_\mathcal{G}$ at every pixel.

\subsection*{Inference}
During inference, we estimate the albedo $A$ and shading $S$ using the probabilistic constrained optimization defined in~\refsec{prob_inference}.
Special care has to be taken when optimizing the shading $S$ factored into gray scale values $B$ and global color $C$.
While the coverall problem is still convex, we found it easier to minimize objective \ref{eq:prob_inference} using an alternating minimization.
We first keep the current estimate for albedo $A$ and $B$ fixed and optimize for a global light color $C$, then keep $A$ and $C$ fixed and optimize for $B$, and finally solve for the albedo $A$ keeping the shading $S=B \cdot C$ fixed.
Each of those steps has a closed form solution.

\begin{table*}
\begin{center}
\resizebox{\linewidth}{!}{%
\begin{tabular}{lc@{\hskip 1mm}c@{\hskip 1mm}c@{\hskip 3mm}c@{\hskip 1mm}c@{\hskip 1mm}c@{\hskip 3mm}c@{\hskip 1mm}c@{\hskip 1mm}c}\\
\toprule
MPI Sintel& \multicolumn{3}{c}{MSE} & \multicolumn{3}{c}{LMSE} & \multicolumn{3}{c}{DSSIM} \\
\cline{2-10}
 & Albedo & Shading & Avg & Albedo & Shading & Avg & Albedo & Shading & Avg\\
\midrule
\multicolumn{10}{l}{Scene Split [Disjoint Training and Testing Scenes]:}\vspace{2mm}\\
L2 loss & 1.93\% & 2.20\% &  2.07\% &  1.22\% &  1.51\% &  1.36\% & 21.22\% & 15.84\% & 18.53\%\\
L2 loss + hard constr. & 3.96\% & 2.21\% &  3.09\% &  2.22\% & 1.52\% & 1.87\% & 20.39\% & 15.84\% & 18.12\% \\
Distr. loss & 1.95\% & 2.07\% & 2.01\% & 1.24\% & 1.43\% & 1.33\% & 22.08\% & 16.44\% & 19.26\%\\
Distr. loss + hard constr. & 3.45\% & 2.07\% & 2.76\% & 2.11\% & 1.43\% & 1.77\% & 19.83\% & 16.28\% & 18.06\%\\
\midrule
Distr. loss + learned constr. & \textbf{1.84\%} & \textbf{1.93\%} & \textbf{1.89\%} & \textbf{1.15\%} & \textbf{1.33\%} & \textbf{1.24\%} & \textbf{17.78\%} & \textbf{13.72\%} & \textbf{15.75\%}\\
\bottomrule
\end{tabular}}
\end{center}
\vspace{-1em}
\caption{Evaluation of different ablations of our model on the MPI Sintel Dataset. We compare our full constrained regression framework with a simple l2 loss with and without hard constraints. Note that all these loss functions are invariant to scale as discussed in~\refsec{details}.}
\lbltbl{sintelAblation}
\end{table*}

\section{Implementation Details}
\lblsec{details}
We evaluate our constrained structured regression framework on the task of intrinsic image decomposition.
Our CNN architecture for modeling the output and constraints distribution is derived from VGG architecture~\citep{simonyan2014very}.
It is similar to VGG from conv1 to conv4, which is then followed by three stride 2 deconvolution to upsample the output to original size. 
Each of these deconvolution layers are followed by rectified linear unit (ReLU) activation function.
We initialize truncated-VGG layers using Imagenet~\citep{imagenet} pretrained weights, and randomly initialize the deconvolution layers.

Notice that albedo and shading for a given image are measured upto a scale, similar to depth.
Thus, we use scale invariant learning procedure~\citep{eigen2014depth}.
We adjust the scale of our prediction by a global constant for each of our output and constraint distribution.
For log domain, it is equivalent to adjusting optimal shift i.e. for output $Y_k$ with target $\tilde{Y}_k$, optimal shift $\alpha$ is computed as follows
\begin{align}
\underset{\alpha}{\text{minimize}}\quad & - \|\alpha+Y_k-\tilde{Y}_k\|_2^2 +\beta\|\alpha\|_2^2\notag,
\end{align}
where $\beta=0.5$ is the regularization coefficient.
Note that regularization $\beta$ is crucial to prevent the network from learning arbitrary scaling.
We shift output as $Y_k=Y_k+\alpha$ and continue the learning procedure as described in~\refsec{technical}.

Our complete implementation is done in caffe~\citep{caffe} and we use ADAM~\citep{kingma2014adam} as stochastic gradient solver.
The learning rate is kept fixed at $1e-4$ throughout the process.
Our method trains in 5K-10K iterations, taking about 4-5 hours using cpu implementation of constraint structured loss.
Complete source code and trained models will be released upon acceptance of publication.

\section{Experiments}
\lblsec{exp}
For our evaluation of intrinsic image decomposition, we use MPI Sintel Dataset~\citep{sintel}.
We use the same setup followed by~\citet{yu:albedo15} and~\citet{chen2013simple}.
Dataset contains total of 890 images with albedo and shading ground truth from 18 scenes.
Following~\citet{yu:albedo15}, we do not train on the image-based split proposed by~\citet{chen2013simple} because there is a large overlap in the image content in training and testing set. 
Instead, we report results on the scene split and report the average error using 2-fold cross-validation, similar to \citet{yu:albedo15}.

We compare all algorithm using the mean squared error (MSE), local mean squared error (LMSE) and structural dissimilarity (DSSIM) metric as proposed by \citet{chen2013simple}.
While both MSE and LMSE measure the raw pixelwise difference between predictions, DSSIM tried to capture the perceptual visual difference between outputs.

\reftbl{sintelAblation} compares different variations of our algorithm.
We start with a baseline L2 loss on our VGG based architecture.
While this L2 loss performs quite well, combining it with inference time constraints significantly degrades the performance.
Our distribution based loss performs similarly to the simple L2 loss, however it absorbs the constraints much better.
Even in this setting constraints actually hurt the raw numeric performance in all settings except for SSIM, meaning that the L2 pixelwise error increased while the perceptual quality of the output slightly increased.
However, as soon as we learn the constraint satisfaction our structured regression framework significantly outperforms the baselines.

\begin{table*}
\begin{center}
\resizebox{\linewidth}{!}{%
\begin{tabular}{lc@{\hskip 1mm}c@{\hskip 1mm}c@{\hskip 3mm}c@{\hskip 1mm}c@{\hskip 1mm}c@{\hskip 3mm}c@{\hskip 1mm}c@{\hskip 1mm}c}\\
\toprule
MPI Sintel& \multicolumn{3}{c}{MSE} & \multicolumn{3}{c}{LMSE} & \multicolumn{3}{c}{DSSIM} \\
\cline{2-10}
 & Albedo & Shading & Avg & Albedo & Shading & Avg & Albedo & Shading & Avg\\
\midrule
\multicolumn{10}{l}{Image Split [Overlapping Training and Testing Scenes]:}\vspace{2mm}\\
\citet{grosse2009ground} &6.06\% &7.27\% &6.67\% &3.66\% &4.19\% &3.93\%&22.70\% &24.00\% &23.35\% \\
\citet{lee2012estimation} & 4.63\% & 5.07\% & 4.85\% &2.24\% &1.92\% &2.08\% &19.90\% &17.70\% &18.80\% \\
\citet{barronTPAMI2015} & 4.20\% & 4.36\% & 4.28\% & 2.98\% & 2.64\% & 2.81\% & 21.00\% & 20.60\% & 20.80\%\\
\citet{chen2013simple} & 3.07\% & 2.77\% & 2.92\% & 1.85\% & 1.90\% & 1.88\% & 19.60\% & 16.50\% & 18.05\%\vspace{2mm}\\
\midrule
\multicolumn{10}{l}{Scene Split [Disjoint Training and Testing Scenes]:}\vspace{2mm}\\
\citet{yu:albedo15} & 2.09\% & 2.21\% & 2.15\% & 1.35\% & 1.44\% & 1.39\% & 20.81\% & 16.08\% & 18.44\%\\
Our Method & \textbf{1.84\%} & \textbf{1.93\%} & \textbf{1.89\%} & \textbf{1.15\%} & \textbf{1.33\%} & \textbf{1.24\%} & \textbf{17.78\%} & \textbf{13.72\%} & \textbf{15.75\%}\\
\midrule
Relative Gain  & 13.59$\%$ & 14.51$\%$ & 13.76$\%$ & 17.39$\%$ & 8.27$\%$ & 12.10$\%$ & 17.04$\%$ & 17.20$\%$ & 17.08$\%$ \\
\bottomrule
\end{tabular}}
\end{center}
\vspace{-1em}
\caption{Comparison of our constrained regression framework with all prior state of the art on MPI Sintel Dataset. Note that our final model leads to a relative improvement between $10\%$ to $20\%$ throughout all metrics.}
\lbltbl{sintelRes}
\end{table*}

Finally, we compare our direct intrinsic regression output to the state-of-the-art intrinsic image decompositions in \reftbl{sintelRes}.
Note that we outperform the best prior work by anywhere between $10\%$ and $20\%$ in relative terms.

\section{Discussion}
We proposed a generic framework for deep structured regression, exploiting inference time constraints.
The constrained structured regression framework is general and easy to train.
The method is potentially applicable to a broad set of application areas e.g., monocular depth estimation~\citet{eigen2014depth}, optical flow prediction~\cite{horn1981determining}, etc.
We illustrated performance on a state of the art intrinsic image decomposition task, where our results confirmed that adding structure back to inference can substantially improve the performance of deep regression models both visually and quantitatively.

\begin{figure*}[t]
\centering
\resizebox{\textwidth}{!}{%
\begin{tabular}{c@{\hskip 1mm}c@{\hskip 1mm}c@{\hskip 0.15in}c@{\hskip 1mm}c@{\hskip 1mm}c}

\rotatebox[origin=lc]{90}{\parbox{0.115\linewidth}{\centering\footnotesize Input Image}} &
\multicolumn{2}{c}{\includegraphics[width=0.27\linewidth]{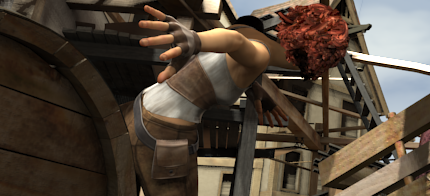}} &
& \multicolumn{2}{c}{\includegraphics[width=0.27\linewidth]{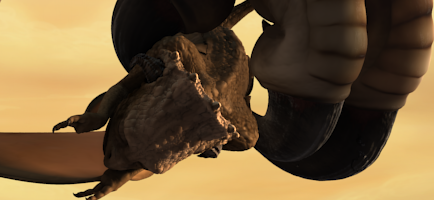}}\\

\rotatebox[origin=lc]{90}{\parbox{0.115\linewidth}{\centering\scriptsize Ground Truth}} &
\includegraphics[width=0.27\linewidth]{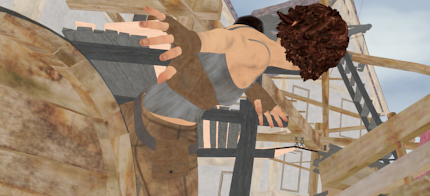} &
\includegraphics[width=0.27\linewidth]{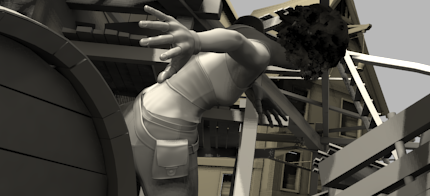} &
\rotatebox[origin=lc]{90}{\parbox{0.115\linewidth}{\centering\scriptsize Ground Truth}} &
\includegraphics[width=0.27\linewidth]{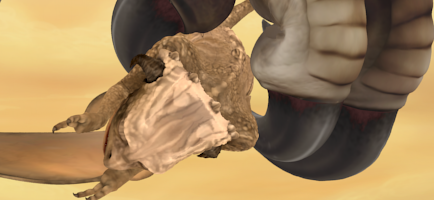} &
\includegraphics[width=0.27\linewidth]{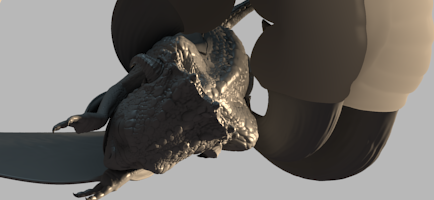}\\

\rotatebox[origin=lc]{90}{\parbox{0.115\linewidth}{\centering\scriptsize\scriptsize Lee et al.}} &
\includegraphics[width=0.27\linewidth]{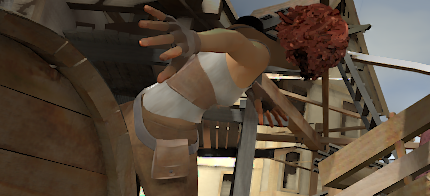} &
\includegraphics[width=0.27\linewidth]{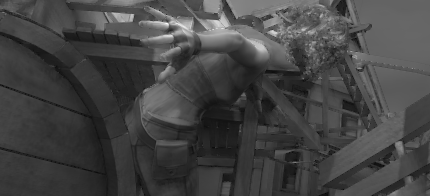} &
\rotatebox[origin=lc]{90}{\parbox{0.115\linewidth}{\centering\scriptsize\scriptsize Lee et al.}} &
\includegraphics[width=0.27\linewidth]{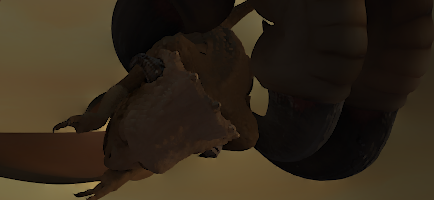} &
\includegraphics[width=0.27\linewidth]{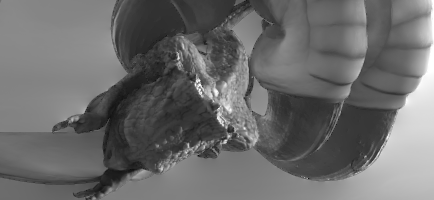}\\

\rotatebox{90}{\parbox{0.115\linewidth}{\centering\scriptsize Barron \& Malik}} &
\includegraphics[width=0.27\linewidth]{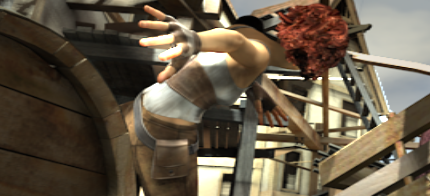} &
\includegraphics[width=0.27\linewidth]{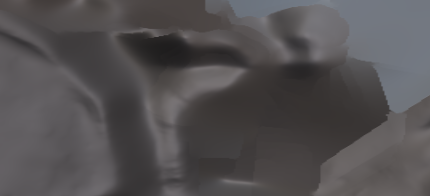} &
\rotatebox{90}{\parbox{0.115\linewidth}{\centering\scriptsize Barron \& Malik}} &
\includegraphics[width=0.27\linewidth]{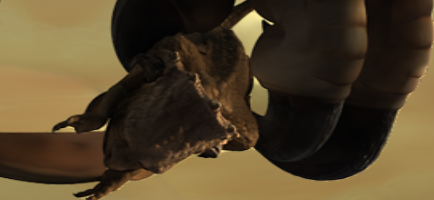} &
\includegraphics[width=0.27\linewidth]{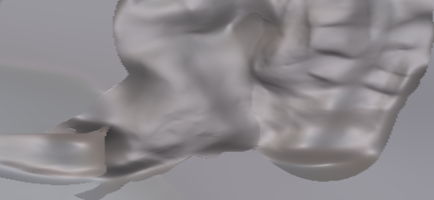}\\

\rotatebox{90}{\parbox{0.115\linewidth}{\centering\scriptsize Chen \& Koltun}} &
\includegraphics[width=0.27\linewidth]{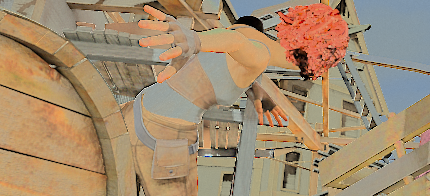} &
\includegraphics[width=0.27\linewidth]{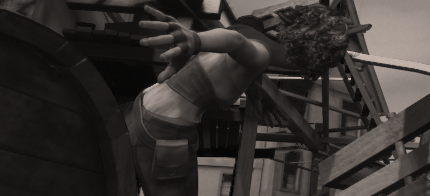} &
\rotatebox{90}{\parbox{0.115\linewidth}{\centering\scriptsize Chen \& Koltun}} &
\includegraphics[width=0.27\linewidth]{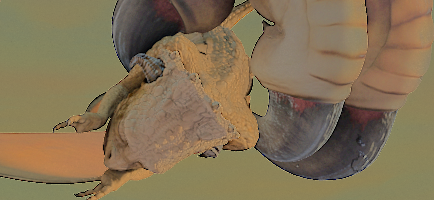} &
\includegraphics[width=0.27\linewidth]{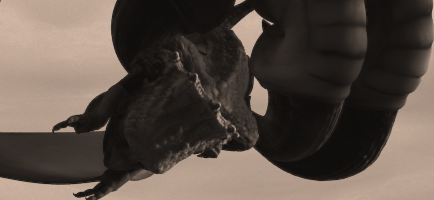}\\

\rotatebox[origin=lc]{90}{\parbox{0.115\linewidth}{\centering\scriptsize Narihira et al.}} &
\includegraphics[width=0.27\linewidth]{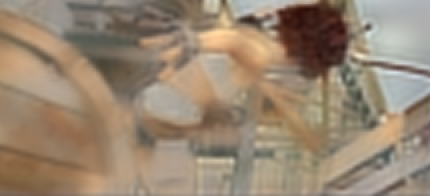} &
\includegraphics[width=0.27\linewidth]{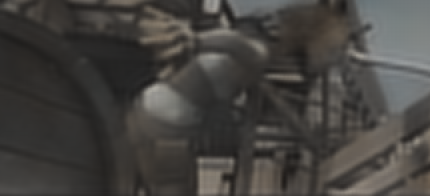} &
\rotatebox[origin=lc]{90}{\parbox{0.115\linewidth}{\centering\scriptsize Narihira et al.}} &
\includegraphics[width=0.27\linewidth]{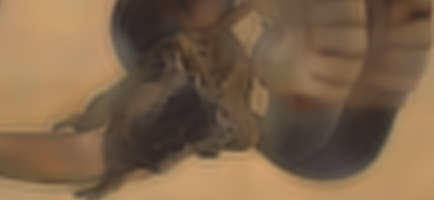} &
\includegraphics[width=0.27\linewidth]{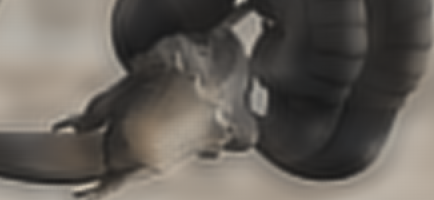}\\

\rotatebox{90}{\parbox{0.115\linewidth}{\centering\scriptsize Ours}} &
\includegraphics[width=0.27\linewidth]{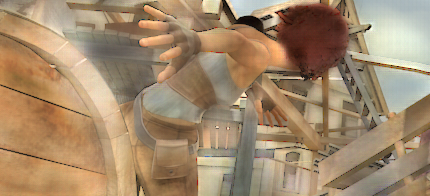} &
\includegraphics[width=0.27\linewidth]{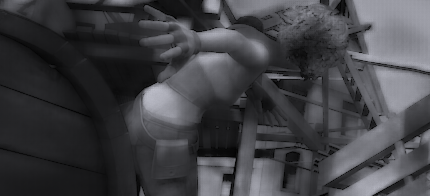} &
\rotatebox{90}{\parbox{0.115\linewidth}{\centering\scriptsize Ours}} &
\includegraphics[width=0.27\linewidth]{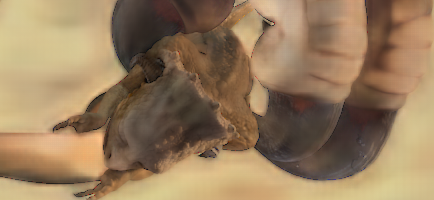} &
\includegraphics[width=0.27\linewidth]{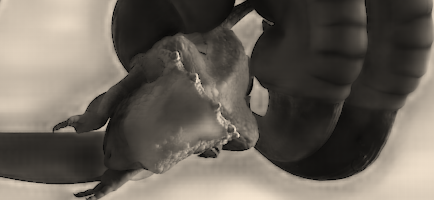}\\

&{\footnotesize Albedo} & {\footnotesize Shading} && {\footnotesize Albedo} & {\footnotesize Shading}\\
\end{tabular}}
\caption{Qualtiative comparision of the results on MPI Sintel Dataset. Note that all approaches, except ours and \citet{yu:albedo15}, uses depth image as input. Our approach outperforms all other approaches significantly just using rgb image. Our results are not only correct, but also very detailed contrary to \citet{yu:albedo15}.}
\lblfig{finalRes}
\end{figure*}

\bibliographystyle{iclr2016_conference}
\bibliography{main}

\end{document}